%% file: negativeStepSize.tex
\documentclass{opt2024} 

\usepackage{natbib}
\usepackage{color}
\usepackage{booktabs}       

\input{math_commands.tex}

\definecolor{red}{rgb}{1,0,0}

\definecolor{gry}{rgb}{0.5,0.5,0.5}
\def\gry#1{{\color{gry}#1}}
\def\norm#1{\|#1\|}

\title[Negative Step Sizes]{Don't Be So Positive: Negative Step Sizes in Second-Order Methods}

\optauthor{%
\Name{Betty Shea} \Email{sheaws@cs.ubc.ca}\\
\Name{Mark Schmidt$^1$} \Email{schmidtm@cs.ubc.ca}\\
\addr The University of British Columbia, Vancouver, Canada.
$^1$Canada CIFAR AI Chair (Amii)}

\begin{document}

\maketitle

\begin{abstract}%
The value of second-order methods lies in the use of curvature information. Yet, this information is costly to extract and once obtained, standard methods to achieve global convergence often discard valuable negative curvature information. This limits the effectiveness of second-order methods in modern machine learning. In this paper, we show that second-order and second-order-like methods are promising optimizers for neural networks provided that we add one ingredient: \emph{negative step sizes}. We show that under very general conditions, methods that produce ascent directions are globally convergent when combined with a Wolfe line search that allows both positive and negative step sizes. We experimentally demonstrate that using negative step sizes is often more effective than common Hessian modification methods. 
\end{abstract}


\section{Introduction}
\label{sec:introduction}
Training neural networks involves minimizing a non-convex objective. Often, second-order methods are considered unsuitable for this task because they are attracted to saddle points and local maxima. By and large, gradient based methods such as gradient descent (GD) with momentum, Adam \citep{Kingma2014} and RMSprop \citep{Tieleman2012} are seen as the only viable optimizers in machine learning. This is a source of frustration. Newton's method and second-order-like methods, such as limited-memory quasi-Newton (QN) methods, excel in traditional machine learning but can fail to even converge in the deep learning setting. First-order methods converge but slowly, as good convergence progress can only be expected if curvature information of the loss landscape is used. We want optimizers that use negative curvature information but these methods may yield search directions that point the wrong way. 

\begin{figure}
\label{fig:2dNegNewtonExample}
\begin{center}
	\includegraphics[width=0.49\linewidth]{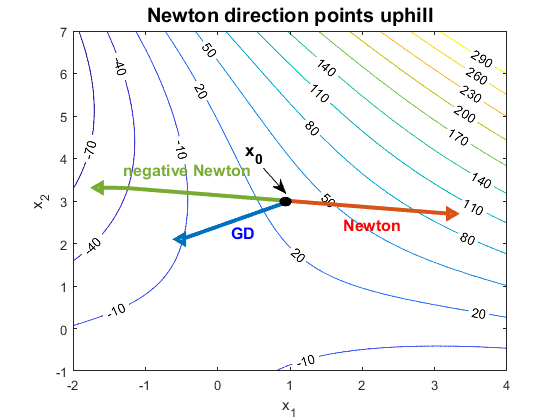}
\end{center}
\caption{In this two-dimensional non-convex minimization problem, the negative of the Newton's direction (green) is a better search direction than the gradient descent direction (blue), Newton's direction (red) or any non-negative combination of both.}
\end{figure}

Many variants of second-order methods ensure global convergence. Popular approaches include Hessian modifications, trust-region methods and cubic regularization (see Section \ref{sec:relatedWork} for a short summary). These methods use the same trick: align the search direction closer to the gradient direction to ensure descent. Yet, this approach could slow down optimization progress. For problems that are ill-conditioned and non-convex, GD ensures progress but at an extremely slow rate \citep{Martens2010}. When the second-order direction points uphill, simply taking the opposite direction may be a better choice than moving the step closer to the gradient step. The negative direction points downhill \emph{and} maintains all second-order information (see Figure \ref{fig:2dNegNewtonExample}). To the best of our knowledge, using negative step sizes is relatively unexplored in optimization. We found one mention of negative step sizes in an analysis of the Fletcher-Reeves \cite{Fletcher1964} non-linear conjugate gradient method \cite{Dai1999} for differentiable and Lipschitz-smooth objective functions. 

\subsection{Contribution}
This paper examines the largely unexplored role of negative step sizes in optimization and shows that taking a backward step is a computationally inexpensive way to incorporate negative curvature information. Our experiments suggest that a quasi-Newton (QN) method, symmetric rank one (SR1) \cite{Conn1991}, combined with negative step sizes, may be an overlooked method useful for training neural networks. This is achieved without using common approaches to ``fix'' second-order methods\footnote{In this paper, we refer to second-order-like methods that approximate the Hessian, such as QN methods, as ``second-order methods''. This is technically incorrect because these methods do not compute second derivatives. This misuse of terminology is for readability.} such as Hessian modifications. We also show that extending the Wolfe line search to both positive and negative step sizes ensures that second-order methods satisfy the Zoutendijk condition. Thus, second-order methods combined with good step size choices are globally convergent even in non-convex settings.

\section{Common approaches for globally convergent second-order methods}
\label{sec:relatedWork}

Two major drawbacks of using second-order methods to train neural networks are (a) the computation and storage cost of the (approximate) Hessian and, (b) non-convergence. To some extent, the first issue is addressed by QN methods and their limited memory versions \cite{Nocedal2006}. In this section, we summarize some common approaches to the second issue of non-convergence.

\paragraph{Hessian modifications} There are many variants of second-order method that modify the (approximate) Hessian for positive definiteness and thus ensure a descent direction (see Sections 3.4 of \cite{Nocedal2006}). A spectral decomposition will find the negative eigenvalues. One could then replace the negative eigenvalues by small positive values, flip the sign of the negative eigenvalues \cite{Paternain2019} or shift the Hessian by subtracting a diagonal matrix that contains the most negative eigenvalue of the Hessian (also known as damping). These modifications, however, require an expensive eigenvalue decomposition and may no longer reflect accurate curvature information. Removing negative eigenvalues could also be detrimental for high-dimensional non-convex problems because following paths of negative curvature is a way to escape saddle points \cite{Choromanska2015,More1979}. 

\paragraph{Trust-region approaches} Trust region (TR) methods are an option for obtaining global convergence and, unlike line searches, they work with methods that produce ascent directions \cite{Conn2000}. Variants of TR methods include the Leveberg-Marquadt \cite{Levenberg1944,Marquadt1963} and dogleg \citep{Powell1970} methods. Newton's and QN methods often employ a TR approach that minimizes the model function over a two-dimensional subspace \cite{Sorensen1982,Byrd1988,Schultz1985,Ramanmurthy2016}. Saddle-free Newton's method \cite{Pascanu2014,Dauphin2014}, Newton's method with \emph{generalized} TR, was proposed as a version of Newton's method that can escape saddle points. However, because the solution to the TR subproblem is equivalent to adding a positive definite diagonal matrix to the (approximate) Hessian \cite{More1983}, this approach also loses curvature information. TR conjugate gradient methods \cite{Steihaug1983} use directions of negative curvature, but this also combines the original search direction with the negative gradient direction. 

\paragraph{Cubic regularization} Cubic-regularization (CR) uses an additional third-order term so that the second-order method is not attracted to local maxima and saddle points. Originally introduced for Newton's method \cite{Nesterov2006}, CR was recently extended to QN methods \citep{Kamzolov2023,Benson2018}. However, CR generally requires special solvers for its third-order subproblem. CR is equivalent to using a Hessian that is modified to be positive definite \cite{Malitsky2020}. 

\paragraph{Positive definite-enforcing updates} Some QN methods, such as BFGS \citep{Broyden1967,Fletcher1964,Goldfarb1970,Shanno1970}, are designed to guarantee positive definiteness of the Hessian approximation at every update step as long as the step size used satisfies the Wolfe conditions \citep{Wolfe1969}. It has been suggested, however, that QN variants that do not enforce positive definiteness, such as SR1, converge faster by adhering closer to the true Hessian \citep{Conn1991,Khalfan1993}. Other updates that enforce positive (semi-) definiteness include natural gradient methods \cite{Amari1998} that use the covariance matrix of the gradients instead of the Hessian, Gauss-Newton methods and Kronecker-factored approximate curvature (KFAC) \cite{Martens2015}. These methods, however, may also sacrifice accurate curvature information by enforcing positive (semi-) definiteness.



\section{Global convergence of ascent-descent methods}
\label{sec:convergence}

Even though negative step sizes may result in a larger decrease on some iterations, we may be concerned that the algorithm might not converge. In this section, we give a general result showing that methods that may produce ascent directions and that allow negative step sizes converge under very general conditions. Our goal is to minimize a function $f:\sR^n\rightarrow\sR$ without constraints. The function $f$ is assumed to be twice differentiable, Lipschitz-smooth and bounded below. We use a deterministic iterative algorithm that, on the $k$th iteration, calculates a search direction $\vp_k$ and step size $\alpha_k$, and updates its iterate $\vx$ by
\begin{equation}
	\label{eq:update_rule}
	\vx_{k+1}=\vx_k+\alpha_k\vp_k.
\end{equation}

Given our iterate $\vx_k$ and a search direction $\vp_k$, we can rewrite $f$ as a function of the step size $\phi(\alpha)=f(\vx_k+\alpha\vp_k)$. The directional derivative is defined as its derivative with respect to $\alpha$, i.e. $\phi'(\alpha)=\nabla f(\vx_k+\alpha\vp_k)^\top\vp_k$. We refer to $\vp_k$ as an \textit{ascent} or a \textit{descent} direction when $\phi'(0)$ is positive or negative respectively. We say that a method is \textit{globally convergent} if it finds a stationary point using update rule \ref{eq:update_rule} starting from any initial estimate $\vx_0$. Further details of notation, definitions and assumptions are given in Appendix \ref{sec:app_notation}.

It is known that methods that produce descent-only directions with step sizes satisfying the Wolfe conditions \cite{Wolfe1969} are globally convergent for a loss function $f$ that satisfies the assumptions above \cite{Nocedal2006}. The proposition below extends this result to methods that produce both ascent and descent directions, and that satisfy a variant of the Wolfe conditions that considers both positive and negative step sizes.  Global convergence holds with the looser requirement that $\vp_k$ is not orthogonal to the gradient direction.

\begin{proposition}
\label{prop:global_convergence}
Consider any algorithm with updates of the form \eqref{eq:update_rule} where $\vp_k$ is a direction that is not orthogonal to the gradient direction, and non-zero step sizes $|\alpha_k|\geq\epsilon>0$ satisfy the Wolfe conditions (\ref{eq:Armijo} and \ref{eq:Wolfe}). Suppose that $f$ is bounded below in $\mathcal{R}^n$ and that $f$ is continuously differentiable in an open set $\mathcal{N}$ containing the level set $\mathcal{L}\triangleq\{\vx:f(\vx)\leq f(\vx_0)\}$, where $\vx_0$ is the initial iterate. Assume also that the gradient $\nabla f$ is Lipschitz continuous on $\mathcal{N}$. Then, the algorithm satisfies Zoutendijk's condition (\ref{eq:zoutendijk}) is therefore globally convergent (\ref{eq:globally_convergent}).
\end{proposition}

The proof of Proposition \ref{prop:global_convergence} is given in Appendix \ref{sec:app_proofProp1}. A modified Wolfe line search that also considers negative step sizes could be easily implemented. We give one possible implementation, which we call \texttt{Wolfe$\pm$}, in Appendix \ref{sec:app_wolfepm}.

\section{Experiments}
\label{sec:experiments}

\begin{figure}
	\centering{
		\includegraphics[width=0.32\linewidth]{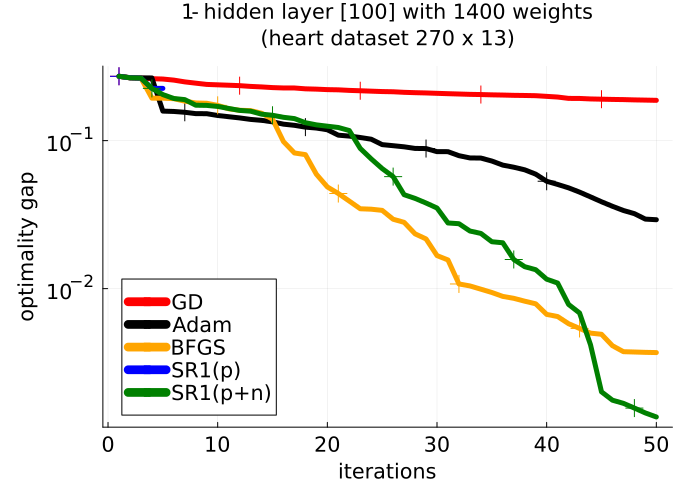}
		\includegraphics[width=0.32\linewidth]{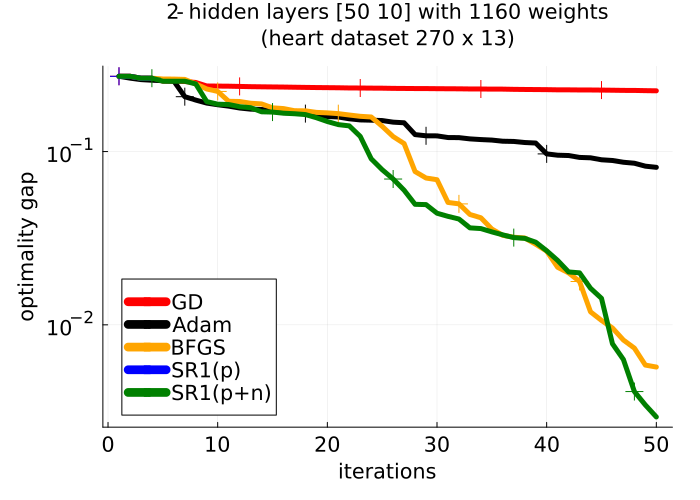}
		\includegraphics[width=0.32\linewidth]{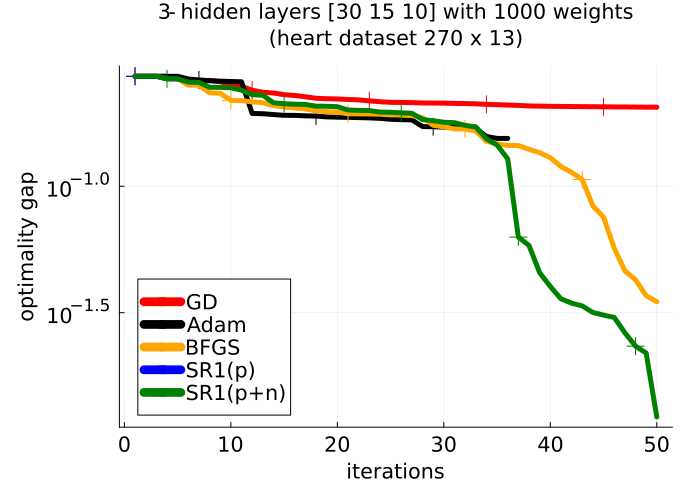}
	}
	\caption{Using negative step sizes in training neural networks. The plots show training error by iteration for neural networks with 1-, 2- and 3- hidden layers on the \texttt{heart} dataset. We compare full QN methods BFGS (yellow) and SR1 (blue and green) with GD (red) and Adam (black). SR1 is non-convergent with positive-only step sizes (blue) but outperforms when step sizes are allowed to be negative (green). The outperformance appears more pronounced in deeper networks. }
	\label{fig:fullQN_trainError}
\end{figure}

\subsection{QN methods and neural network training}
\label{sec:experiments_fullQN}
In this experiment, we looked at full QN methods and their performance on neural network training. All methods use a line search that gives step sizes satisfying the Wolfe conditions (\texttt{p}). SR1 is further combined with \texttt{Wolfe$\pm$} (\texttt{p+n}) that uses negative step sizes when $\vp_k$ is not a descent direction. QN methods are initialized with the identity matrix. Methods are run for a maximum of 50 iterations. Figure \ref{fig:fullQN_trainError} shows training error by iteration. 

Because the objective is non-convex, the true Hessian at iterate $\vx_k$ may not be positive definite. Thus, SR1 produces Hessian approximations that are not positive definite and a $\vp_k$ that does not point downhill. With a regular line search, SR1 diverges within the first few steps (blue). With negative step sizes, SR1 becomes the an effective optimizer. This effect appears to be more pronounced as the number of network layers increase. BFGS combined with a Wolfe line search guarantees positive definiteness in its Hessian update and thus never diverges even with positive only step sizes. Yet, a positive definite Hessian may not accurately capture the optimization landscape. The lack of negative curvature information in the BFGS Hessian approximation may be the reason its training error is higher than that of SR1 after the maximum number of iterations is reached. For further details, please refer to Appendix \ref{sec:app_experiments}.

\subsection{Limited-memory QN methods and neural network training}
\label{sec:experiments_lQN}

\begin{table}
	\begin{center}
		\begin{tabular}{llcccccc}
			\toprule
			Dataset & size & GD & Adam & l-BFGS & l-SR1 & l-SR1+damp & l-SR1+\texttt{Wolfe$\pm$}\\
			\midrule
			a1a & $(1605 \times 119)$ & 0.217 & 0.203 & 0.315 & 0.500 & 0.214 & \textbf{0.199} \\
			a9a & $(32561 \times 123)$ & 0.220 & 0.212 & 0.332 & 0.330 & 0.216 & \textbf{0.211} \\
			colon-cancer & $(62 \times 2000)$ & 0.006 & 0.046 & 0.500 & 0.183 & \gry{\textit{DNF}} & \textbf{0.003}\\
			gisette & $(6000 \times 5000)$ & 0.139 & 0.171 & 0.395 & 0.395 & \gry{\textit{DNF}} & \textbf{0.050}\\
			heart & $(270 \times 13)$ & 0.455 & 0.382 & 0.388 & \textbf{0.237} & 0.239 & \textbf{0.237} \\
			ijcnn1 & $(35000 \times 22)$ & 0.132 & \textbf{0.120} & 0.131 & 0.165 & 0.130 & \textbf{0.120}\\
			ionosphere & $(351\times 34)$ & 0.201 & \textbf{0.118} & 0.201 & 0.500 & 0.199 & 0.170 \\
			leukemia & $(38\times 7129)$ & 0.028 & \textbf{0.001} & 0.160 & 0.160 & \gry{\textit{DNF}} & 0.064\\
			madelon & $(2000\times 500)$ & 0.500 & 0.500 & 0.490 & 0.396 & \textbf{0.382} & 0.396 \\
			mushrooms & $(8124 \times 112)$ & 0.040 & 0.018 & 0.166 & 0.500 & 0.019 & \textbf{0.009} \\
			splice & $(1000\times 60)$ & 0.270 & 0.260 & 0.485 & 0.500 & 0.262 & \textbf{0.195}\\
			svmguide3 & $(1243\times 22)$ & 0.295 & 0.276 & 0.374 & 0.374 & 0.267 & \textbf{0.260} \\
			w1a & $(2477\times 300)$ & 0.088 & \textbf{0.059} & 0.200 & 0.200 & 0.071 & 0.063\\
			w8a & $(49749\times 300)$ & 0.096 & \textbf{0.066} & 0.193 & 0.193 & 0.079 & 0.071\\
			\bottomrule
		\end{tabular}
		\caption{Comparing the use of negative step sizes with damping, a common Hessian modification, for limited-memory QN methods. The experiment was run across several datasets fitted with a neural network. The lowest training error achieved in each dataset is highlighted in bold. Limited-memory SR1 with positive only step sizes (\texttt{l-SR1}) often does not converge. Damping (\texttt{l-SR1+damp}) helps with convergence but is an expensive operation and in many cases did not finish training in the allocated time (shown as \textit{DNF}). Using negative step sizes (\texttt{l-SR1+Wolfe$\pm$}) showed good performance even against a state-of-the-art optimizer such as Adam. }
		\label{tbl:QNexperiments}
	\end{center}
\end{table}

Table \ref{tbl:QNexperiments} shows the result of using two limited-memory QN methods, l-BFGS \cite{Liu1989} and l-SR1 \cite{Byrd1994}, to train a neural network across different datasets. Gradient descent (GD) and Adam are plotted for comparison. The purpose of this experiment is to compare two globalization strategies: damping and negative step sizes. Damping requires an eigenvalue decomposition and is therefore an expensive operation. Thus, in many cases where the dataset has a large number of features and the Hessian is large (e.g. \texttt{colon-cancer}), l-SR1 combined with damping did not finish in the allocated amount of time (shown as \textit{DNF}). l-BFGS combined with a Wolfe line search does not require damping because its Hessian approximations are guaranteed to be positive definite. Yet, a lack of negative curvature information may hurt its effectiveness in training neural networks, and it tends to perform worse than l-SR1 with negative step sizes.

\section*{Acknowledgements}

Betty Shea is funded by an NSERC Canada Graduate Scholarship. The work was partially supported by the Canada CIFAR AI Chair Program and NSERC Discovery Grant RGPIN-2022-036669.

\bibliography{biblio}
\pagebreak
\appendix
\section{Notation and further background details}
\label{sec:app_notation}

\noindent We say an algorithm is \emph{globally convergent} if it produces a sequence of gradients that converge to zero, or
\begin{equation}
\label{eq:globally_convergent}
\liminf_{k\rightarrow\infty}\norm{\nabla f(\vx_k)}=0.
\end{equation}
\\
\noindent The angle $\theta_k$ between search direction $\vp_k$ and the steepest descent direction $-\nabla f(\vx_k)$ is given by
\begin{equation}
\label{eq:cosine_similarity}
\cos\theta_k=\frac{-\nabla f(\vx_k)^\top\vp_k}{\norm{\nabla f(\vx_k)}\norm{\vp_k}}.
\end{equation}
Equation (\ref{eq:cosine_similarity}) is sometimes referred to as the cosine similarity between the $\vp_k$ and the steepest descent direction. Search direction $\vp_k$ is not orthogonal to the gradient if $|\cos\theta_k|\geq\delta>0$.\\
\\
Zoutendijk's condition is satisfied if 
\begin{equation}
\label{eq:zoutendijk}
\sum_{k\geq0}\cos^2\theta_k\norm{\nabla f(\vx_k)}^2<\infty,
\end{equation}
which implies that
\[\cos^2\theta_k\norm{\nabla f(\vx_k)}^2\rightarrow0.\]
\\
\noindent If $f$ has a \emph{Lipschitz continuous gradient} on open set $\mathcal{N}$, then there exists a constant $L>0$ such that
\begin{equation}
\label{eq:Lipschitz_gradient}
\norm{\nabla f(\vx)-\nabla f(\tilde{\vx})}\leq L\norm{\vx-\tilde{\vx}}\text{, for all }\vx,\tilde{\vx}\in\mathcal{N}
\end{equation}
\\
\noindent A line search is an auxiliary method that finds a step size $\alpha_k$ along $\vp_k$ that has desirable properties. For example, for constants $c_1\in(0,1)$ and $c_2\in(c_1,1)$ and descent direction $\vp_k$, a step size $\alpha_k>0$ that satisfies the Wolfe condition has two desirable properties. Firstly, $\alpha_k$ guarantees making progress in minimizing $f$, or
\begin{equation}
\label{eq:Armijo}
f(\vx_k+\alpha_k\vp_k)\leq f(\vx_k)+c_1\alpha_k\nabla f(\vx_k)^\top\vp_k.
\end{equation}
Equation \ref{eq:Armijo} is also known as the Armijo condition. Secondly, $\alpha_k$ does not lie too far from an optimal step size choice.
\begin{equation}
\label{eq:Wolfe}
0\geq\nabla f(\vx_k+\alpha_k\vp_k)^\top\vp_k\geq c_2\nabla f(\vx_k)^\top\vp_k.
\end{equation}
Equation \eqref{eq:Wolfe} is also known as the curvature condition. Taken together, equations \eqref{eq:Armijo} and \eqref{eq:Wolfe} are often referred to as the Wolfe conditions. Note that this standard curvature condition assumes that $\vp_k$ is a descent direction and thus $\nabla f(\vx_k)^\top\vp_k<0$. In our case where $\vp_k$ is an ascent direction, the curvature condition becomes
\begin{equation}
\label{eq:Wolfe_ascent}
0\leq\nabla f(\vx_k+\alpha_k\vp_k)^\top\vp_k\leq c_2\nabla f(\vx_k)^\top\vp_k.
\end{equation} 


\section{Proof of Proposition 1}
\label{sec:app_proofProp1}
\begin{proof}
The proof of Theorem \ref{prop:global_convergence} follows closely that of Theorem 3.2 in \cite{Nocedal2006}, which considers only the case of descent directions paired with positive step sizes. Here we extend the results to ascent directions $\vp_k$ paired with negative step sizes $\alpha_k$.\\
\\
\noindent Combining update rule (\ref{eq:update_rule}) and curvature condition (\ref{eq:Wolfe_ascent}) where $\nabla f(\vx_k)^\top\vp_k>0$ and $\alpha_k<0$ gives
\begin{equation}
\label{eq:prop1_proof_a}
\left(\nabla f(\vx_k)-\nabla f(\vx_{k+1})\right)^\top\vp_k\geq (1-c_2)\nabla f(\vx_k)^\top\vp_k.
\end{equation}


\noindent Lipschitz continuous gradient gives
\[\norm{\nabla f(\vx_k)-\nabla f(\vx_{k+1})}\leq L\norm{\vx_k-\vx_{k+1}}=L\norm{-\alpha_k\vp_k}=|-\alpha_k|L\norm{\vp_k}=-\alpha_kL\norm{\vp_k}\]
and
\begin{equation}
\label{eq:prop1_proof_b}
\left(\nabla f(\vx_{k})-\nabla f(\vx_{k+1})\right)^\top\vp_k\leq-\alpha_k L\norm{\vp_k}^2.
\end{equation}
Combining (\ref{eq:prop1_proof_a}) and (\ref{eq:prop1_proof_b})
\[(1-c_2)\nabla f(\vx_k)^\top\vp_k\leq-\alpha_kL\norm{\vp_k}^2.\]
Rearranging gives
\begin{equation}
\label{eq:prop1_proof_c}
\alpha_k\leq-\frac{1-c_2}{L}\frac{\nabla f(\vx_k)^\top\vp_k}{\norm{\vp_k}^2}\leq0
\end{equation}
where the middle term in inequality  (\ref{eq:prop1_proof_c}) is negative because $0<c_1<c_2<1$ and $\nabla f(\vx_k)^\top\vp_k>0$. Using (\ref{eq:prop1_proof_c}) with the sufficient decrease condition (\ref{eq:Armijo}) gives
\begin{align*}
	f(\vx_{k+1})&\leq f(\vx_k)+c_1\alpha_k\nabla f(\vx_k)^\top\vp_k\\
	&\leq f(\vx_k)-\frac{c_1(1-c_2)}{L}\frac{\left(\nabla f(\vx_k)^\top\vp_k\right)^2}{\norm{\vp_k}^2}\\
	&=f(\vx_k)-\frac{c_1(1-c_2)}{L}\left(\frac{\left(\nabla f(\vx_k)^\top\vp_k\right)}{\norm{\nabla f(\vx_k)}\norm{\vp_k}}\right)^2\norm{\nabla f(\vx_k)}^2\\
	&=f(\vx_k)-c\cdot\cos^2\theta_k\norm{\nabla f(\vx_k)}^2
\end{align*}
where the last step uses (\ref{eq:cosine_similarity}) and $c=\frac{c_1(1-c_2)}{L}$. Summing across iterations 0 to $k$,
\[f(\vx_{k+1})\leq f(\vx_0)-c\sum_{j=0}^k\cos^2\theta_j\norm{\nabla f(\vx_j)}^2\]
Because $f$ is bounded below, we know that $f(\vx_0)-f(x_k)$ is less than a positive constant for all $k$. Taking limits gives
\[\sum_{k=0}^\infty\cos^2\theta_k\norm{\nabla f(\vx_k)}^2<\infty\]
and thus satisfies the Zoutendijk condition, and implies that 
\[\cos^\theta_k\norm{\nabla f(\vx_k)}^2\rightarrow0\]
Because $|\cos\theta_k|\geq\delta>0$ for all $k$, then $\cos^2\theta_k\geq\delta^2>0$ for all $k$. This further implies that 
\[\lim_{k\rightarrow\infty}\norm{\nabla f(\vx_k)}=0\]
and thus the algorithm is globally convergent.
\end{proof}

\section{Generalized Wolfe conditions}
\label{sec:app_wolfepm}

Suppose we have an implementation of the standard Wolfe line search (\cite{Nocedal2006} Section 3.5). Then \texttt{Wolfe$\pm$} can be implemented trivially as in Algorithm \ref{alg:Wolfepm}. First we check the sign of the directional derivative. If $\vp_k$ is a descent direction, then we continue by calling the standard Wolfe line search. If $\vp_k$ is an ascent direction, then the standard Wolfe line search is called with $-\vp_k$ as the direction and the step size returned is negated afterwards. 

\begin{algorithm}
	\label{alg:Wolfepm}
	\SetAlgoLined
	\KwIn{$\alpha_{\max}>0$, $p_k$, function \texttt{Wolfe($p_k$,$\alpha_{\max}$)} }
	\KwOut{step size $\alpha_{W\pm}$ satisfying \texttt{Wolfe$\pm$}}
	\eIf {$\nabla f(x_k)^\top p_k<0$}{
		$\alpha_{W\pm}\leftarrow$ \texttt{Wolfe($p_k$,$\alpha_{\max}$)};}
	{
		$\alpha_{W\pm}\leftarrow-$ \texttt{Wolfe($-p_k$,$\alpha_{\max}$)};}
	\caption{\texttt{Wolfe$\pm$} is easy to implement with access to a standard Wolfe line search.}
\end{algorithm}

\section{Further details on experiments}
\label{sec:app_experiments}

\begin{figure}
\centering{
	\includegraphics[width=0.32\linewidth]{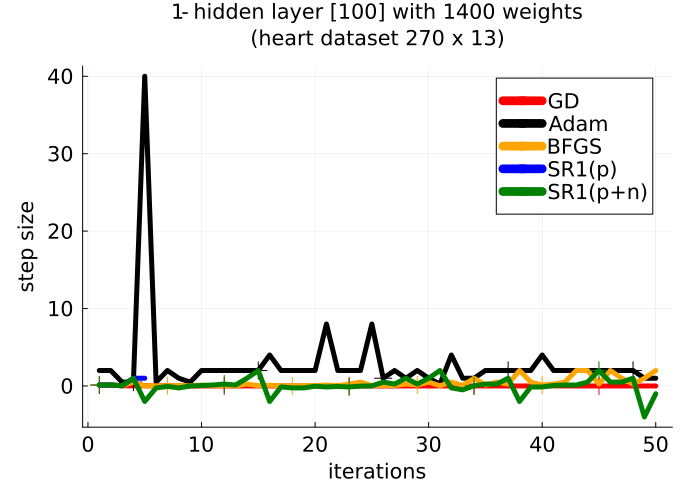}
	\includegraphics[width=0.32\linewidth]{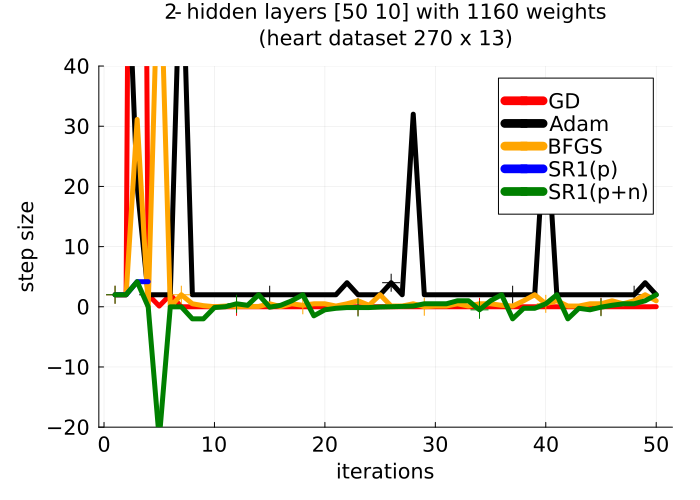}
	\includegraphics[width=0.32\linewidth]{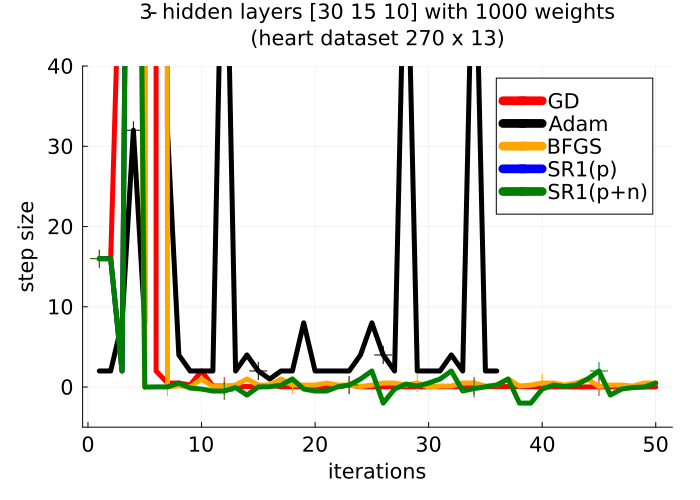}
	\includegraphics[width=0.32\linewidth]{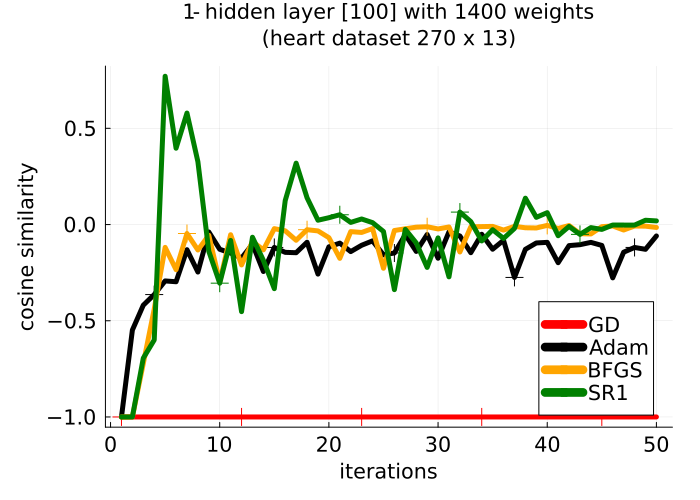}
	\includegraphics[width=0.32\linewidth]{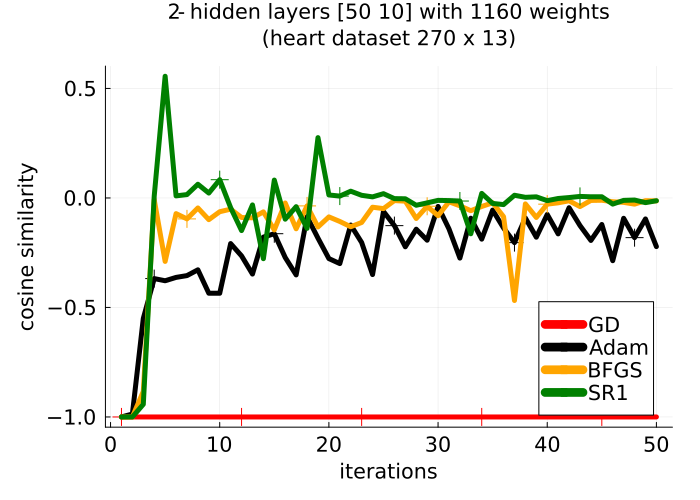}
	\includegraphics[width=0.32\linewidth]{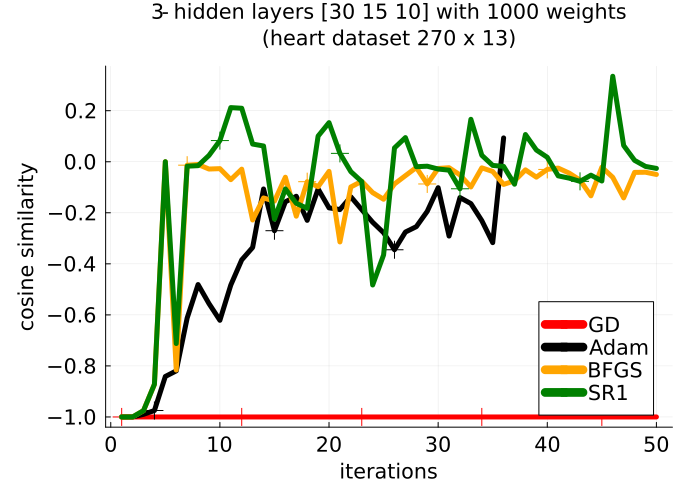}
}
\caption{Step sizes (top row) and cosine similarity (bottom row) of different optimizers for neural networks with 1, 2 and 3 hidden layers. SR1 often produces ascent directions and this is dealt with effectively by using negative step sizes.}
\label{fig:fullQN_stepsize_cosSim}
\end{figure}

Figure \ref{fig:fullQN_stepsize_cosSim} plots the step sizes and cosine similarity of the search direction $\vp_k$ with the steepest descent direction of the different optimizers in the experiments discussed in Section \ref{sec:experiments_fullQN}.

\end{document}

%% file: math_commands.tex
\usepackage{amsmath,amsfonts,bm}

\def\eqref#1{equation~\ref{#1}}

\def\1{\bm{1}}

\def\vp{{\bm{p}}}

\def\vx{{\bm{x}}}

\DeclareMathAlphabet{\mathsfit}{\encodingdefault}{\sfdefault}{m}{sl}
\SetMathAlphabet{\mathsfit}{bold}{\encodingdefault}{\sfdefault}{bx}{n}

\def\sR{{\mathbb{R}}}

